%% file: 0205.tex
\documentclass[runningheads]{llncs}
\usepackage{graphicx}
\usepackage{placeins}
\usepackage{amsmath,amssymb} 
\usepackage{color}
\usepackage{amsmath,amssymb} 
\usepackage{caption}
\usepackage{algorithm}
\usepackage[noend]{algpseudocode}
\usepackage{xspace}
\usepackage{array}
\usepackage{float}
\usepackage{mathtools, cuted}
\usepackage[breaklinks=true,bookmarks=false]{hyperref}

\input{commands}

\usepackage{graphicx}
\usepackage{subcaption}

\begin{document}

\title{Geometric Image Synthesis} 
\titlerunning{Geometric Image Synthesis}


\author{Hassan Abu Alhaija$^1$, Siva Karthik Mustikovela$^1$,\\ Andreas Geiger$^{2,3}$, Carsten Rother$^1$}

%
\index{Abu Alhaija, Hassan}
\index{Mustikovela, Siva Karthik}

\authorrunning{Abu Alhaija et al.} 


\institute{Visual Learning Lab, Heidelberg University, Heidelberg, Germany 
\email{hassan.abu\_alhaija@iwr.uni-heidelberg.de}\\
\and
Autonomous Vision Group, MPI for Intelligent Systems, T\"ubingen, Germany
\and University of T\"ubingen, T\"ubingen, Germany
}

\maketitle

\input{sec00_abstract}
\input{sec01_introduction}
\input{sec02_relwork}
\input{sec03_method}

\input{sec04_experiments}

\input{sec05_conclusion}


\bibliographystyle{splncs04}
\bibliography{bibliography_short,bibliography,bibliography_custom}
\end{document}

%% file: commands.tex
\newcommand{\figref}[1]{\Fig~\ref{#1}}

\renewcommand{\eqref}[1]{Eq.~\ref{#1}}

\makeatletter
\newcommand{\raisemath}[1]{\mathpalette{\raisem@th{#1}}}
\newcommand{\raisem@th}[3]{\raisebox{#1}{$#2#3$}}
\makeatother

\makeatletter
\DeclareRobustCommand\onedot{\futurelet\@let@token\@onedot}
\def\@onedot{\ifx\@let@token.\else.\null\fi\xspace}
\def\eg{e.g\onedot}

\def\etc{etc\onedot} 

\def\etal{et~al\onedot} 
\def\Fig{Fig\onedot}   
\makeatother

\definecolor{darkgreen}{rgb}{0,0.7,0}

%% file: sec00_abstract.tex
\begin{abstract}
The task of generating natural images from 3D scenes has been a long standing goal in computer graphics. On the other hand, recent developments in deep neural networks allow for trainable models that can produce natural-looking images with little or no knowledge about the scene structure. While the generated images often consist of realistic looking local patterns, the overall structure of the generated images is often inconsistent. In this work we propose a trainable, geometry-aware image generation method that leverages various types of scene information, including geometry and segmentation, to create realistic looking natural images that match the desired scene structure. Our geometrically-consistent image synthesis method is a deep neural network, called Geometry to Image Synthesis (GIS) framework, which retains the advantages of a trainable method, e.g., differentiability and adaptiveness, but, at the same time, makes a step towards the generalizability, control and quality output of modern graphics rendering engines. We utilize the GIS framework to insert vehicles in outdoor driving scenes, as well as to generate novel views of objects from the Linemod dataset. We qualitatively show that our network is able to generalize beyond the training set to novel scene geometries, object shapes and segmentations. Furthermore, we quantitatively show that the GIS framework can be used to synthesize large amounts of training data which proves beneficial for training instance segmentation models. 
\end{abstract}

%% file: sec01_introduction.tex
\section{Introduction}

\begin{figure}[t!]
    \centering
    \begin{subfigure}[b]{0.24\textwidth}
        \includegraphics[trim=50 0 50 0,clip,width=\textwidth]{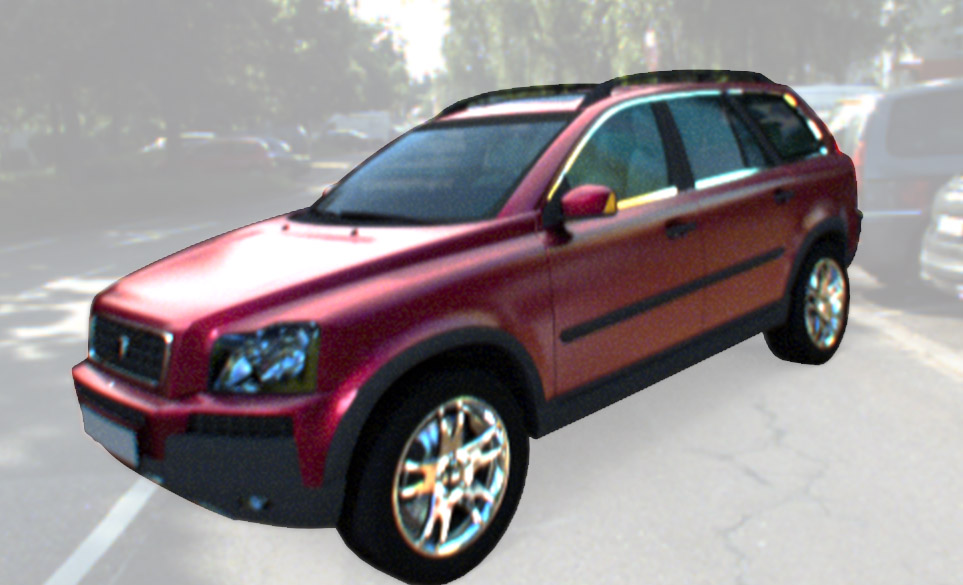}
    \end{subfigure}
    \begin{subfigure}[b]{0.24\textwidth}
        \includegraphics[trim=50 0 50 0,clip,width=\textwidth]{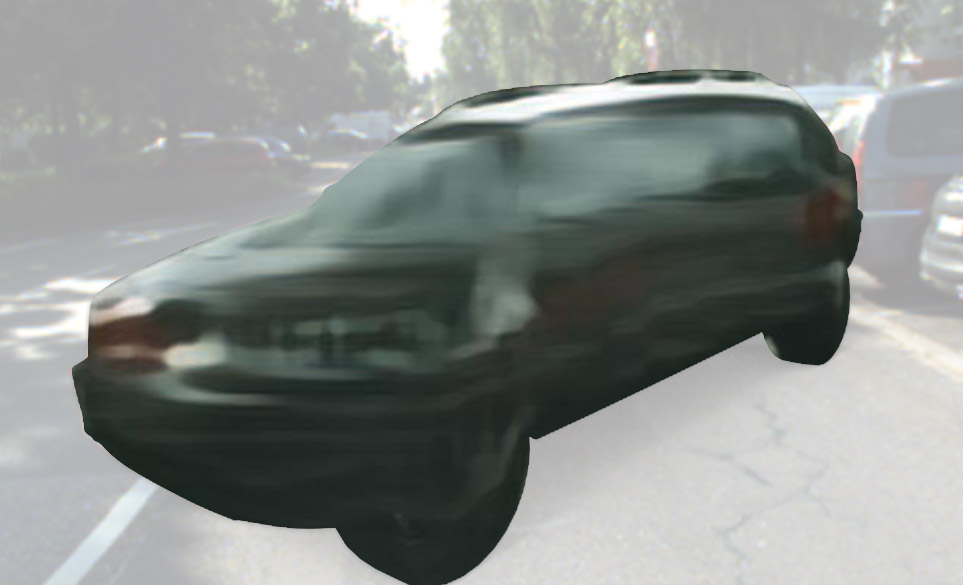}
    \end{subfigure}
    \begin{subfigure}[b]{0.24\textwidth}
        \includegraphics[trim=50 0 50 0,clip,width=\textwidth]{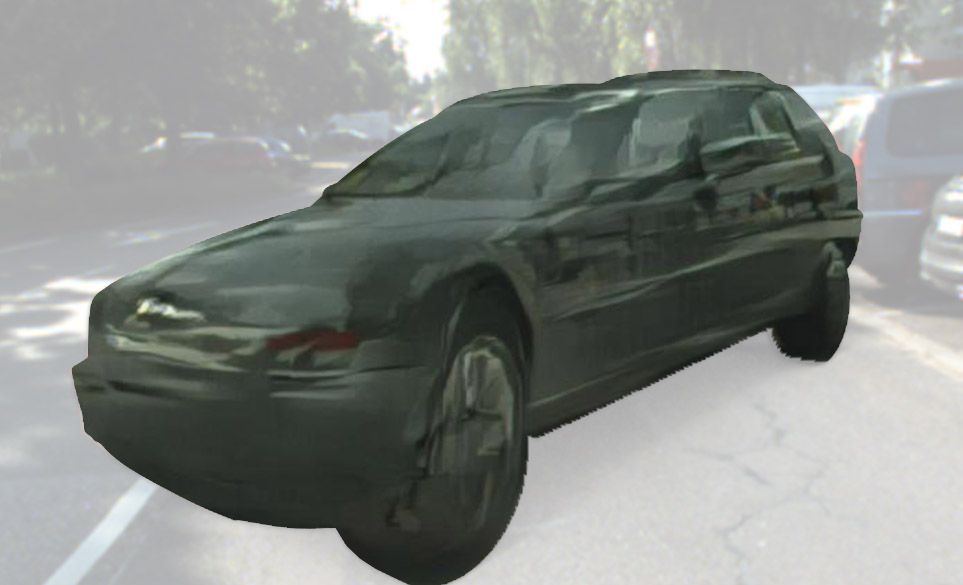}
    \end{subfigure}
    \begin{subfigure}[b]{0.24\textwidth}
        \includegraphics[trim=50 0 50 0,clip,width=\textwidth]{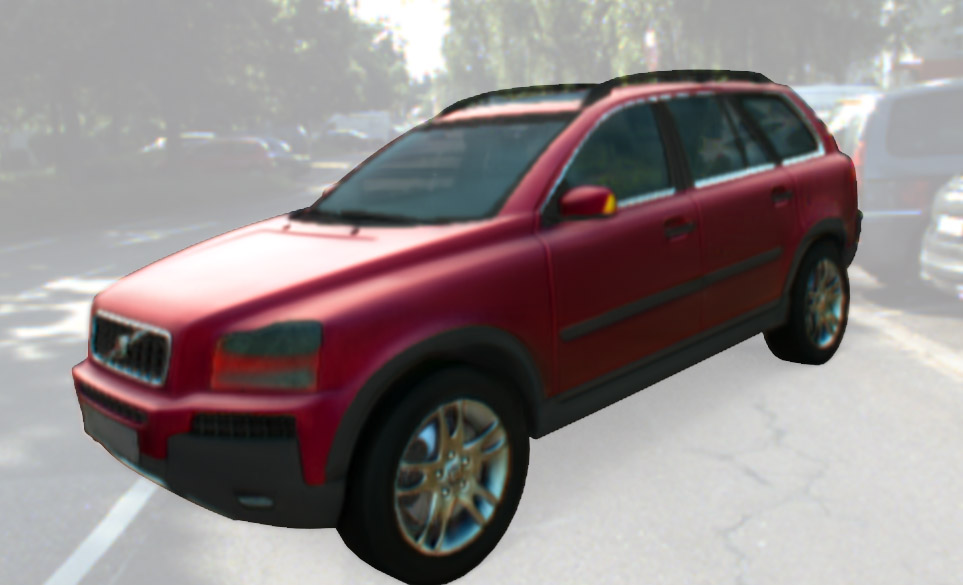}
    \end{subfigure}
\\ 
    \begin{subfigure}[b]{0.24\textwidth}
        \includegraphics[trim=50 0 50 0,clip,width=\textwidth]{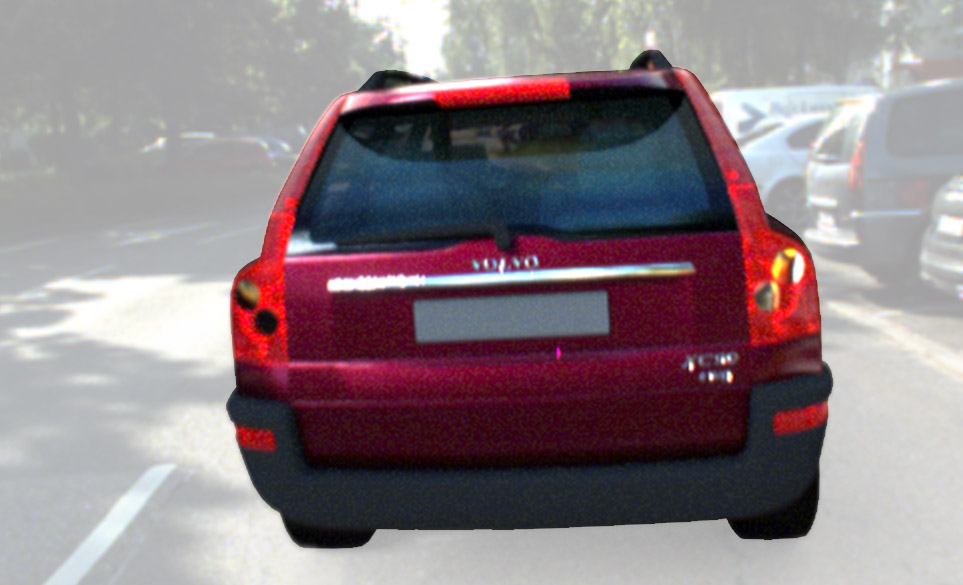}
        \caption{Cycles renderer}
        \label{fig:compare_software}
    \end{subfigure}
    \begin{subfigure}[b]{0.24\textwidth}
        \includegraphics[trim=50 0 50 0,clip,width=\textwidth]{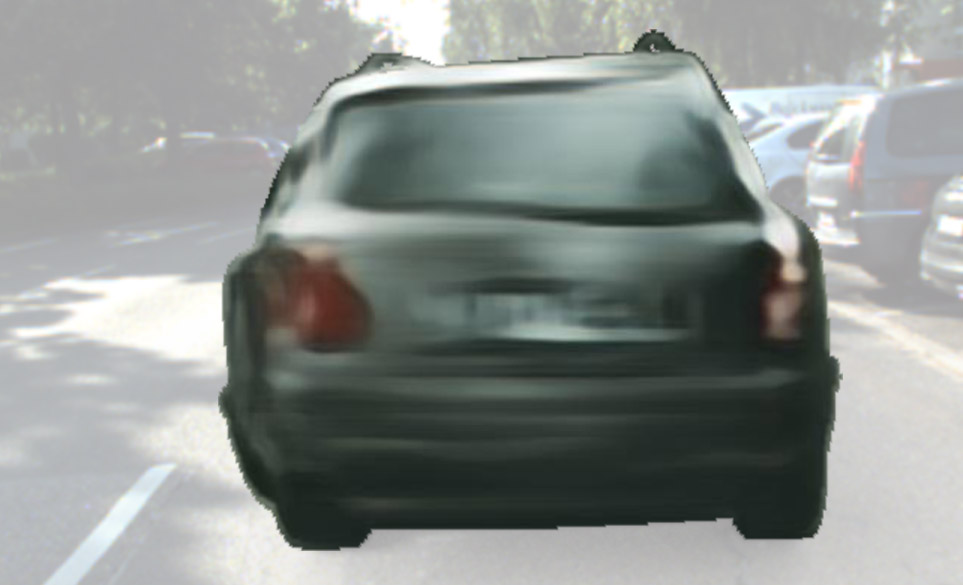}
        \caption{CRN \cite{Chen2017ICCV}}
        \label{fig:compare_crn}
    \end{subfigure}
    \begin{subfigure}[b]{0.24\textwidth}
        \includegraphics[trim=50 0 50 0,clip,width=\textwidth]{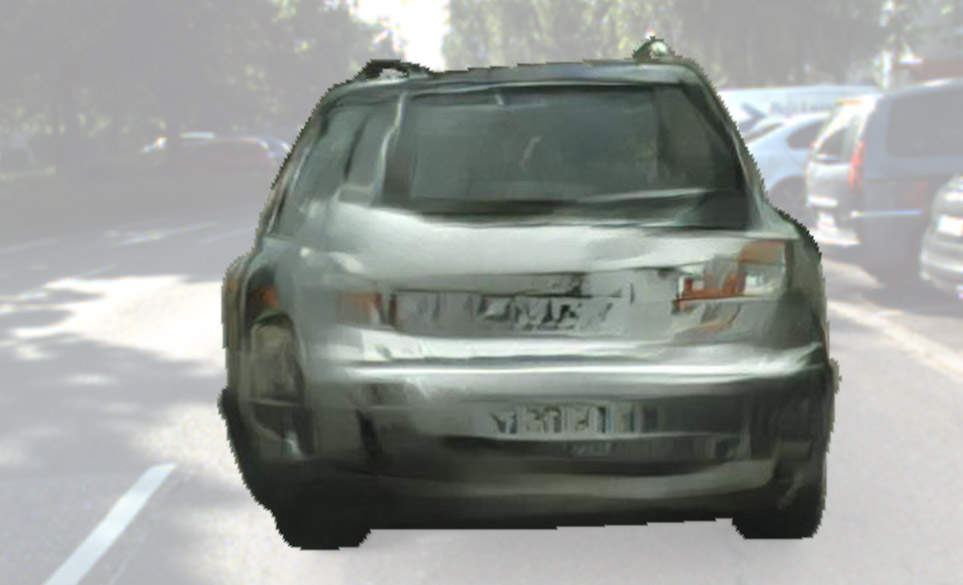}
        \caption{Pix2PixHD \cite{wang2018pix2pixHD}}
        \label{fig:compare_imagetoimage}
    \end{subfigure}
    \begin{subfigure}[b]{0.24\textwidth} 
        \includegraphics[trim=50 0 50 0,clip,width=\textwidth]{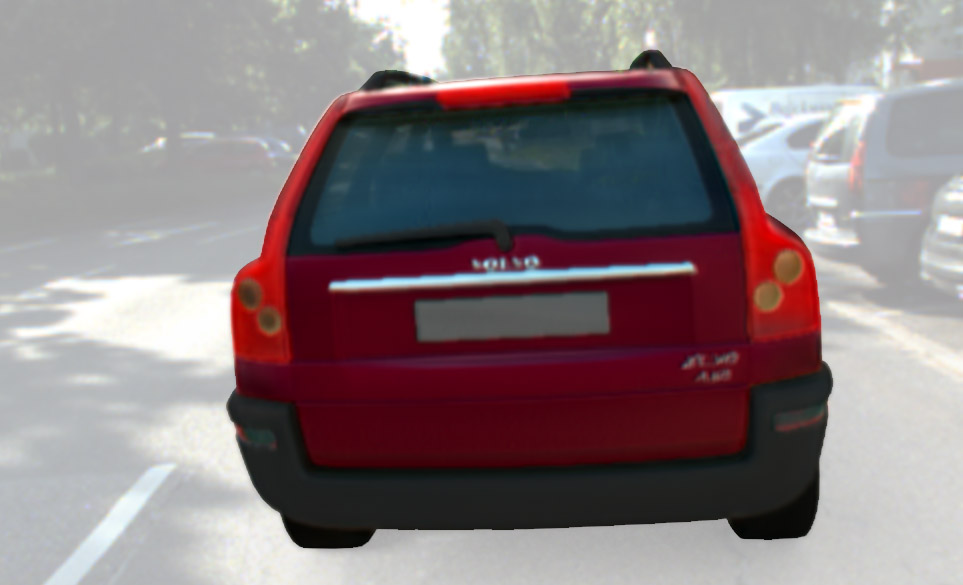}
        \caption{GIS (Ours)}
        \label{fig:compare_GIS}
    \end{subfigure}
    \caption{(a) The result of a state-of-the-art Physically-based Renderer ("Cycle" Renderer).  (b,c) Results of two other deep neural network based image generation methods \cite{Chen2017ICCV},\cite{wang2018pix2pixHD}. While in both cases local image patches looks plausible the whole image does not look realistic. (d) Our GIS framework can realistically synthesize the car object with a specific pose using a deep neural network.}
    \label{fig:compare}
\end{figure}

Methods for generating natural images from noise or sparse input have gained significant interest in recent years with the developments in Generative Deep Neural Networks. Specifically, Generative Adversarial Networks (GANs)\cite{Goodfellow2014NIPS}, allowed for trainable models that can produce natural-looking images with little or no prior knowledge input just by learning to imitate the distribution in a target image set. While the generated images often consist of realistic looking local patterns, the overall structure of the images can be inconsistent. Using more sparse cues, like edge maps or semantic segmentation\cite{Isola2017CVPR}, introduces some local control over the output but doesn't address the global structure. Further, recent works have addressed the problem of global consistency by generating the image at different scales \cite{Chen2017ICCV} or using two separate global and local networks \cite{wang2018pix2pixHD}. These solutions, nevertheless, address the global 2D structure of the image but not the 3D structure of the scene. This is evident when trying to generate an object in a different pose than those present most commonly in the training dataset (see figure \ref{fig:compare_crn}, \ref{fig:compare_imagetoimage}).
While image generation from semantic segmentation can produce visually impressive images, it is not clear whether it can produce new training data for other vision tasks. This could be attributed to two factors: (i) The sparse input makes the image generation problem largely under-constrained leading to inconsistent image structure; (ii) The lack of control parameters over the image generation process (\eg Pose and color of objects) makes it hard to define the desired attributes of the output image.

On the other hand, generating natural images from known 3D geometry, texture and material properties through rendering engines has been widely used to generate training data for various computer vision tasks. While physically-based rendering engines aim at accurately reproducing the physical properties of light-material interactions, most available rendering engines use a set of carefully designed approximations, in order to reduce the computational complexity and produce results that are visually appealing to humans. Rendered images accurately matches the input scene structure but differ in local appearance from real images due the disparity between the real capturing process and the approximations in the software rendering pipeline. Previous works \cite{tremblay2018training} pointed to the performance gap between synthetic and real data when used for training a task like semantic or instance segmentation. 
The other limitation of rendering engines is that they require accurate and complete information about the objects and the scene, namely, detailed 3D geometry, texture and material properties, lighting information, environment maps, and so on. This usually requires laborious manual work by experts to set up the 3D scene.  

In this work, we propose a geometry-aware image generation method that leverages various types of scene information, like geometry and segmentation, to create realistic images which match the desired scene structure and properties. The network is trained with two objectives, the first is a supervised loss where the goal is to learn a mapping from the multi-channel input to an RGB image that matches the input structure. The second is an adversarial loss that learns to compare generated and real images enforcing the generator results to look similar to real data. We explore different input modalities like normals, depth, semantic and material segmentation and compare their usefulness. Using this rich input we are able to show visually a clear improvement over existing state-of-the-art image generation approaches, e.g. \cite{Chen2017ICCV} \cite{Isola2017CVPR} \cite{wang2018pix2pixHD}.

The goal of our approach is not only to generate visually realistic images, but also to explore whether the images generated can be useful for training other networks for various computer vision tasks. The advantage of using a trainable model instead of a software rendering engine is two-fold. 
Firstly, it can produce realistic looking images from geometry and segmentation while learning, from training data, to implicitly predict the remaining rendering parameters (e.g. material properties and lighting conditions).
Secondly, the trainable model has the advantage of producing images that are fine-tuned to specific characteristics of the training dataset by leveraging the Adversarial loss. For instance, it can capture the specific noise distribution and color shifts in the data. 

In order to demonstrate the abilities of our GIS framework, we perform two types of experiments. In the first, we utilize an augmented reality dataset where synthetic vehicles were realistically rendered into a scene using ``Cycles renderer" from Blender \cite{Alhaija2017BMVC}. We use the normals, depth and material labels as input and the rendered images as the target in the supervised loss, while using real car images to train the discriminator in the adversarial loss.  In this way, our network is able to generate realistic looking images (see Fig. \ref{fig:compare_GIS} and supplementary video\footnote[1]{\url{https://youtu.be/W2tFCz9xJoU}}) similar to the rendered data from  \cite{Alhaija2017BMVC} (see fig. \ref{fig:compare_software}). In fact, we train the GIS network to give $9$ diverse output-images and observed that each image captures a different lighting condition (e.g. direct sunshine, clear sky, or cloudy), all present in the training data. Using our trained network, we produce a new dataset of $4000$ augmented images of car objects on top of real driving images. This dataset is used to train a state-of-the-art instance segmentation network, here Mask R-CNN \cite{He2018MaskRCNN}. This improves the performance of Mask R-CNN over the original augmented data \cite{Alhaija2018IJCV}.
In the second experiment, we demonstrate how our network can be trained directly using real images only. For that we utilize the Linemod dataset \cite{hinterstoisser2012accv} that includes images of several objects and their 3D scanned models in addition to the corresponding 6D pose of the objects in each image. We show that using our GIS Network we are able to generate large amount of training data that helps improve the performance of instance segmentation. To summarize, our {\bf contributions} are as follows :
\begin{itemize}
\item We introduce a trainable deep neural network, called GIS, that is able to generate geometry-consistent images from limited input information, like normals and material segmentation, While the remaining aspects of the image, e.g. lighting conditions, are learned from training data.  
\item We qualitatively show that our framework generalizes to novel scene geometries, objects and segmentation, for both synthetic and real data. 
\item We quantitatively show that our network can synthesize training data that improves the performance of a state-of-the-art instance segmentation approach, here Mask R-CNN (\cite{He2018MaskRCNN}). To the best of our knowledge, this is the first time that synthesized training data from a Neural Network is used to advance a state-of-the-art instance segmentation approach.
\end{itemize}

%% file: sec02_relwork.tex
\section{Related work}

\subsubsection{Synthetic Datasets.}

The success of supervised deep learning models has fueled the demand for large annotated datasets. An alternative to tedious manual annotation is provided by the creation of synthetic content, either via manual 3D  scene modeling \cite{Ros2016CVPR,Zhang2017CVPR} or using some stochastic scene generation process \cite{McCormac2017ICCV,Souza2016ARXIV,Tsirikoglou2017ARXIV,Veeravasarapu2016ARXIV}. Mayer \etal \cite{Mayer2018ARXIV,Mayer2016CVPR} demonstrate that simple synthetic datasets with ``flying 3D things'' can be used for training stereo and optical flow models. Ros \etal \cite{Ros2016CVPR} proposed the SYNTHIA dataset with pixel-level semantic segmentation of urban scenes. In contrast, Gaidon \etal \cite{Gaidon2016CVPR} propose ``Virtual KITTI'', a synthetic dataset reproducing in detail the popular KITTI dataset \cite{Geiger2012CVPR}. Richter \etal \cite{Richter2016ECCV,Richter2017ICCV} and Johnson-Roberson \etal \cite{Johnson-Roberson2017ICRA} have been the first to demonstrate that content from commercial video games can be accessed for collecting semantic segmentation, optical flow and object detection ground truth.

An alternative to synthesizing the entire image content is to render only specific objects into natural images. The simplest approach is to cut object instances from one image and paste them onto random background images \cite{Dwibedi2017ICCV} using appropriate blending or GAN-based refinement
\cite{Xu2018ARXIVa}. More variability can be obtained when rendering entire 3D CAD models into the image. Several works consider the augmentation of images with virtual \cite{Cheung2017ARXIV,Hattori2015CVPR} or scanned humans \cite{Chen2016THREEDV,Varol2017CVPR}. In contrast, Abu Alhaija \etal \cite{Alhaija2017BMVC,Alhaija2018IJCV} consider the problem of augmenting scenes from the KITTI dataset with virtual vehicles for improving object detectors and instance segmentation algorithms. In particular, they have shown that a well performing instance segmentation method,  here MNC \cite{Dai2016CVPR}, can be considerably improved by intelligently generating additional training data. 

While great progress has been made in rendering photo-realistic scenes, creating the required content and modeling all physical processes (\eg, interaction of light) correctly is a non-trivial and time-consuming task. In contrast to classical rendering, we propose a generative feed-forward model which maps an intermediate representation of the scene to the desired output. The geometry and appearance cue of this intermediate representation are easily obtained using fast standard OpenGL rendering. 

\subsubsection{Conditional Adversarial Learning.}

Recently, generative adversarial networks (GANs) \cite{Goodfellow2014NIPS} have been proven to be powerful tools for image generation. Isola \etal \cite{Isola2017CVPR} formulate the image-to-image translation problem by conditioning GANs on images from another domain and combining an adversarial with a reconstruction loss. 
Yang \etal \cite{Yang2017ARXIVa} introduce an additional diversity loss to generate more diverse outputs.
Wang \etal \cite{Wang2017ARXIV} propose a multi-scale conditional GAN architecture for generating images of up to 2 Megapixel resolution.
Wang \etal \cite{Wang2016ECCV} use a GAN to synthesize surface normals and another GAN to generate an image from the resulting normal map. GANs' major advantage is that they don't require matching source and target images but rather enforce the generator to produce images that match the target data distribution. We exploit this by adding an Adverserial loss in our GIS framework such that the generated images are realistic. Besides, we explore a richer set of input modalities compared to just raw images \cite{Denton2015NIPS,Wang2016ECCV,Radford2015ARXIV} or semantic segmentations \cite{Isola2017CVPR,Yang2017ARXIVa,wang2018pix2pixHD} for generating higher-quality outputs. We demonstrate that our model compares favorably to the High-Resolution Image Synthesis model of Wang \etal \cite{wang2018pix2pixHD} (see fig \ref{fig:compare_GIS}).

\subsubsection{Feed-Forward Image Synthesis.}

Dosovitskiy \etal \cite{Dosovitskiy2017PAMI} consider an alternative formulation to GANs using feedforward synthesis with a regression loss. Their work demonstrates that an adversarial loss is not necessary to generate accurate images of 3D models given a model ID and a viewpoint. In the same spirit, Chen \etal \cite{Chen2017ICCV} consider the problem of synthesizing photographic images conditioned on semantic layouts using a purely feedforward approach. They demonstrate detailed reconstructions at resolutions up to 2 Megapixels, improving considerable upon the results of Isola \etal \cite{Isola2017CVPR}.

Our work also uses a feedforward formulation for the image synthesis problem. Unlike \cite{Chen2017ICCV}, however, our focus is on synthesizing controllable, high quality images. Thus, we consider 3D geometry and segmentation (semantic or material) as input, provided by a simple OpenGL rendering unit. 

%% file: sec03_method.tex
\section{Method}

A general image generation process can be defined as a mapping $\mathcal{F} : \{G,A,E\} \rightarrow I$ from scene description $\{G,A,E\}$ to an RGB image $I$. The scene description consists of three parts, (i) the geometry parameters $G$ which include the poses and shapes of objects, (ii) the appearance parameters $A$ which describe the objects' materials, textures and transparency and finally (iii) the environment parameters $E$ which describe global conditions of the scene that affect all objects such as lighting, camera parameters, and the environment.  
In this work in contrast, our goal is to train a mapping $\mathcal{R} : \{G,S\} \rightarrow I$ that can produce natural images from a given geometry $G$ and material segmentation $S$ only, without the knowledge of exact appearance $A$ or environment parameters $E$. Similar to semantic segmentation, the material segmentation labels each pixel with a specific material label (\eg metal, glass \etc) from a pre-defined set of materials without providing any properties or parameters of the material. The task of the network is to learn the unknown parameters from the training data directly and apply them to generate images from new input geometry. \\
The target image $I$, which is used for training, can either be a real image of a known scene geometry or a rendered synthetic image obtained through a high-quality software renderer.
While learning image generation directly from real images is desirable, it is often difficult to obtain geometry and material labels which are pixel-accurately aligned with real-world images. For this reason, it is possible to exploit synthetically rendered data using a state-of-the-art physically based renderer as supervised target while using an adversarial loss with real images to acquire realistic looking results.   
Using realistically rendered images also gives us fine-grained control over the data, which we exploit to conduct various experiments for analyzing our model.
Additionally, we demonstrate how our method can be trained directly using real images for the supervised loss when an exact 6D pose of the objects in the image is known. \\

\begin{figure}[t!]
\centering
    \includegraphics[width=\textwidth]{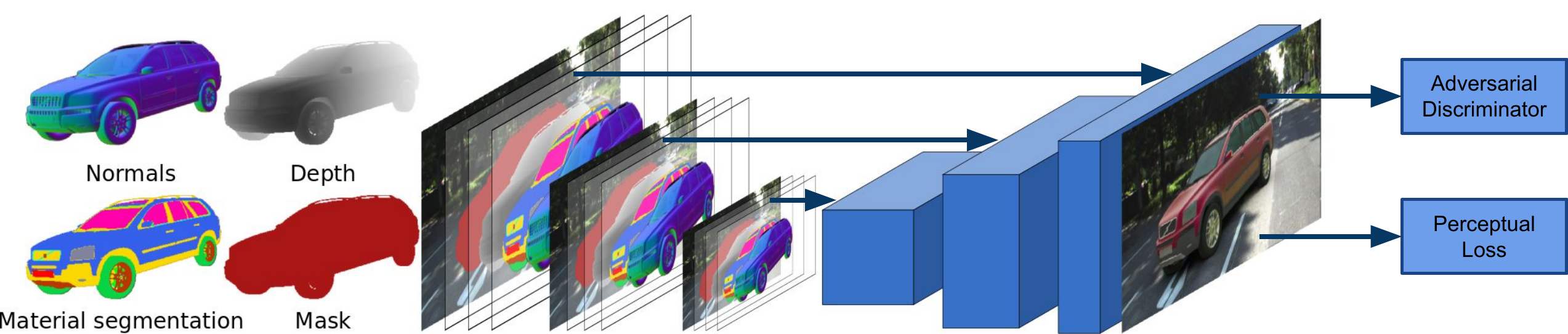}
    \caption{{\bf Overview of our approach.} We propose Geometric Image Synthesis framework, feed-forward architecture for synthesizing RGB images based on geometric and semantic cues. Here we show the case where a car is augmented onto an empty road. Compared to existing image synthesis approaches, our model benefits from a rich set of input modalities, while learning realistic mappings which generalize to novel geometries and segmentations, and integrate the objects seamlessly into provided image content.}
\label{fig:architecture}
\end{figure}

\subsection{Input Representation}\label{subsec:method_data}

Geometry plays a major role in defining the appearance of an object in an image since it defines its shape in addition to its shading through interaction with light. Providing the geometry as an input changes the learning objective from learning to create objects to learning a correlation between geometry and appearance. This makes the network more generalizable to new geometries as we show in later expriments. To use geometry in a deep neural network, it is important to find a compact representation of the object's 3D surface. While meshes are one of the most common representations for 3D objects, they are problematic in the context of convolutional neural networks due to their irregular 3D structure. Another popular representation of 3D objects are voxels. Voxel-based representations can be handled using 3D convolutions \cite{Riegler2017CVPR} but suffer from two shortcomings: high computational requirements and comparably low resolution.

A common 2D representation of 3D shapes is their depth in the camera view. The advantage of such an image-based geometry representation is that it can be directly processed with a regular 2D convolutional neural network. Nevertheless, the object appearance doesn't usually depend on its absolute depth except for secondary effects like lens blur and environmental distortions. Rather, it depends on the small changes in depth between neighbouring points which defines the relative surface orientation with respect to the light source. This can be better characterized by computing the surface normals in the camera coordinates system at each point of the visible surface.

The main advantage of learning the image generation from geometry compared to rendering is the ability to exploit high-level semantic and context cues to predict the appearance of an object. This allows it to {\it learn} non-geometric attributes of the object appearance such as material parameters, lighting, environment reflections and texture directly from data. 
Using semantic \cite{Chen2017ICCV} or instance segmentation \cite{wang2018pix2pixHD} can help the network to learn the appearance of semantically similar objects across multiple examples. Nevertheless, it can be challenging in cases where a semantic class has a large variety in appearance, \eg cars with different models and colors. 
We propose in this work to use material segmentation instead. 
Each pixel in the segmentation input gets a label from a pre-defined set of materials (\eg metal, plastic, glass \etc). 
This doesn't include any material properties or parameters. Rather, it groups parts made of similar materials together allowing the network to learn the material appearance model from multiple objects in different contexts, \eg different lighting conditions. 
This results in more generalization power since the material appearance is often independent of the object class, pose or shape.  We expect this labeling to be particularly effective when generating objects that consist of a small number of materials but vary significantly in shape, \eg, cars.

\subsection{Network Architecture}\label{subsec:method_architecture}
We now define our network architecture in detail.
As discussed before, our goal is to learn a mapping from an intermediate representation to a natural RGB image using a deep neural network.
As input layers to our network, we use the normal map, the depth map, object mask and material segmentation of the object which can be easily obtained using OpenGL based rendering. Additionally, by providing the network with a background image $I_{bg}$, the network can learn to augment synthetic objects realistically into real images \eg add shadows underneath a synthetic car and blending edges.

Fig. \ref{fig:architecture} illustrates the input layers to our network. Let $N,D,S$ be the 2D images representing the normal map, depth map and semantic segmentation of the input object respectively.  Let $M$ denote the material label where each pixel is represented by a one-hot encoding vector which identifies its material ID, see \figref{fig:architecture} for an illustration.
We are now ready to formally represent the mapping as $\mathcal{R} : \{N, D, S, M, I_{bg}\} \rightarrow I$.

For our generator $R$, we follow Chen \etal \cite{Chen2017ICCV} and use a feed-forward coarse-to-fine network architecture for image synthesis. More specifically, we leverage a cascade of convolutional layer modules $C$ starting from a very low resolution input and growing to modules of higher resolutions  (Fig. \ref{fig:architecture}). Each convolutional module $C_i$ has an input resolution of $w_i \times h_i$ and produces a feature map $F_i$ of the same size. $C_i$ receives the feature map $F_{i-1}$ from the previous module, upsampled to $w_i \times h_i$, and concatenated with the input downscaled to the same resolution. 
The following layer, $C_{i+1}$, operates at twice the resolution of the previous layer ($2w_i \times 2h_i$), and receives the feature maps $F_{i}$ and the input rescaled to $2w_i\times 2h_i$. 
Each convolutional module $C_i$ consists of an input layer, intermediate layer and output layer, each of which is followed by a set of convolutions, a layer normalization and a leaky ReLU nonlinearity. The output layer of the final module is followed by a $1\times 1$ convolution applied to the feature map and normalized to obtain the synthesized image. 
For the adversarial discriminator $D$, we adopt a fully convolutional network architecture consisting of 5 convolutional layers each followed by a leaky ReLU with a stride of 2 for all except the last layer. The discriminator's output is a 2D binary map where each value describes the discriminator classification of a patch as real or synthesized by the generator. This is specially useful when synthesizing objects into real images where the same image would contain both real and synthetic patches.  To further stabilize the adversarial training, we employ the simple discriminator gradient regularization method proposed in \cite{roth2017stabilizing}

\subsection{Training}\label{subsec:method_training}
 
We train the generator $R$ in our GIS framework to produce synthesized images $I_s$ that resemble the target images $I_t$ obtained using the ``Cycles'' rendering engine while at the same time being close in appearance to real images in order to ``confuse'' the adversarial discriminator. Effectively, the task of the network is to learn the process of generating images, directly from the target images, given $\{N, D, S, M, I_{bg}\}$ without information such as lighting, environment map or material properties. Those properties are estimated by the network during training and combined with the geometry and segmentation input to produce a high quality image. To achieve this, we choose to compute the perceptual loss (feature matching loss) as proposed in \cite{Gatys2015ARXIV} between the generated and target image. The goal of the perceptual loss is to match the feature activations produced by $I_t$ and $I_s$ at various convolutional levels through a perception network, \eg, VGG. This helps the network to learn fine-grained image patterns while also preserving the global object structure. We use VGG pre-trained on ImageNet as our perceptual network. Let us denote this network by $V$, and let $V_l$ denote a layer of this network. The loss between $I_t$ and $I_s$ is given by 
$$
\mathcal{L}^{P} = \sum\limits_{l} S_l \, \lambda_l \, {\Vert V_l(I_t) - V_l(I_s) \Vert}_1
$$
where $V_l(\cdot)$ denotes the feature activations of VGG at layer $l$ and $S_l$ is the binary mask of the object rescaled to the size of $V_l$.
The GIS framework can also learn to synthesize objects on top of real images. In this case, our goal is to create augmented images by learning not just the target object appearance but also its interaction with the environment in the real image, including shadows, reflection and blending at the object's edges.  
Towards this goal, we add the background image to the input and train the network using an $\ell_1$ loss for the background areas outside the object mask
$
\mathcal{L}^{B} = (1-S){\Vert I_t - I_s \Vert}_1 . 
$
The adversarial discriminator $D_s$ is trained to segment the augmented images by the generator into real and synthetic parts using the Binary Cross-Entropy loss
$
\mathcal{L}^{D} = \mathbb{E}[\log( S - D_s(I_s) )].
$
By replace the synthetic object mask $S$ with its inverse $(1-S)$, we define an adversarial loss for the generator  
$
\mathcal{L}^{A} = \mathbb{E}[\log( (1-S) - D_s(I_s) )]
$
that evaluates the realism of the synthesized objects.
%

\subsection{Diversity}\label{subsec:method_diversity}
Synthesizing images from geometry and segmentation alone is an ill-posed problem. That is, for a specific set of inputs, there are infinite plausible outputs due to different possible lighting conditions, object colors etc. Thus, we task our network to produce $K$ diverse outputs from the last layer using multiple-choice learning \cite{Guzman-Rivera2012NIPS,Chen2017ICCV}. More specifically, we compute the loss for each of these outputs, but only back propagate the gradient of the best configuration for the foreground prediction, while averaging the background predictions as none of them should deviate from the input:
$
\mathcal{L} = w\min_{k} [\,\mathcal{L}_k^{P} + \mathcal{L}_k^{A}] + (1-w)\,\frac{1}{K}\sum_k\mathcal{L}_k^{B} .
$
where $w$ is a weight inversely proportional to the number of pixels of the synthesized object(s).
Note that only the foreground object with the smallest loss is taken into account, thus the min operator effectively acts as a multiple choice switch. This encourages the network to output a diverse set of images to spread its bets over the domain of possible images that could be produced from the current input. In all our experiments we use $K=9$ as the number of diverse outputs, see Fig. \ref{fig:diversity} for an illustration.

%% file: sec04_experiments.tex
\section{Experiments}
To demonstrate the ability of our GIS network to synthesize realistic images, we perform a set of experiments which assess the quality and generalization capacity of the method. We mainly focus on two scenarios, outdoor driving and indoor objects. Realistically synthesizing augmented objects like cars or obstacles into real-world scenes is an important feature for expanding manually annotated training datasets. 
In the case of indoor objects, a learned network can be used to synthesize novel views of objects to provide extensive training data for various tasks. In the following experiments, we show that our GIS framework produces better images for training an instance segmentation network compared to a state-of-the-art software rendering engine. 

\subsection{Augmentation of KITTI-360 dataset}

Introduced by Abu Alhaija \etal \cite{Alhaija2017BMVC}, the augmented KITTI-360 dataset features 4000 augmented images obtained from 200 real-world images through carefully rendering up to 5 high-quality 3D car models into each image using classical rendering techniques. The set of 28 car models have been manually created and placed to ensure high realism. Rendering was performed using the physically-based Cycles Renderer in Blender and followed by a manually designed post-processing pipeline to increase the realism of the output.
Additional scene information like 360 degree environment maps and camera calibration has been used to ensure realistic reflections and good integration with the real image.


To train the GIS network, we use normals, depth, material segmentation and semantic masks of the augmented cars obtained using the augmented KITTI-360 pipeline. The material labels include 16 materials of different properties (\eg plastic, chrome glass etc.) in addition to 15 car paint materials which differ only in color. We use the corresponding RGB images from augmented KITTI-360 as target images for training the parameters of GIS network. Mixing the real images with rendered cars presents an additional challenge since interactions between the inserted objects and the real background, \eg shadows and transparencies, have to be taken into account. To deal with this, we input the background image to the network in addition to geometry and segmentation.  The network's task is to learn the process of synthesizing cars realistically and blend them into the surrounding environment by appropriately adding reflections, shadows and transparencies, amongst others. 

During inference, we obtain a new set of car model positions and orientations following the procedure in \cite{Alhaija2017BMVC}. We render the mask, depth, normals and material labels from the camera viewpoint and use them as input to our GIS network. 
Note that during the inference phase, we do not require a sophisticated rendering pipeline, like ``Cycles'' renderer, since normals, depth maps and segmentations can be obtained directly using a simple OpenGL based renderer. We then leverage the trained GIS model to create a new dataset of 4000 augmented images with new car poses and combinations.  

\subsubsection{Qualitative evaluation:}

\begin{figure}[t!]
\centering
\includegraphics[width=\textwidth]{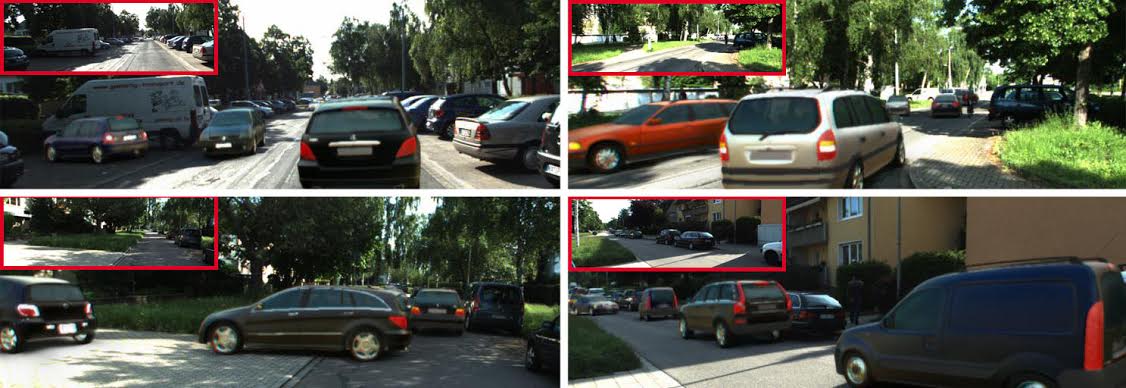}
\caption{Images from KITTI-360 dataset augmented with cars synthesized using our GIS framework (Real image without augmentation in upper left corner).
}
\label{fig:results_kitti360}
\end{figure}

\figref{fig:results_kitti360} shows augmented images produced by our GIS framework when trained on the augmented KITTI-360 dataset. Note that the synthesized cars exhibit realistic appearance properties like shading, shadows, reflections and specularity, despite the fact that this information is not provided to our model. The material labeling of the cars allows the model to tune the synthesis process to each material. Importantly, note that the material label is just a semantic label of material and does not contain any information with respect to the physical properties of the material. Interestingly, our model is able to learn the transparency property of the material with label "glass" from data, without providing any alpha channel or explicitly modeling transparency. Additionally, the model is able to replicate camera effects such as blur and chromatic aberrations, which are present in the augmented KITTI-360 dataset.

\subsubsection{Quantitative evaluation:}

To verify the effectiveness of data produced by our model, we train the state-of-the-art Mask R-CNN model \cite{He2018MaskRCNN} for car instance segmentation using the images produced by our network. Alongside, we also train the same model with images from the original Cycles Rendering pipeline from \cite{Alhaija2017BMVC} with the same 3D car models and poses, and a baseline model using the unaugmented real images from KITTI-360. We evaluate all models on the KITTI 2015 training set. The results are presented in Table \ref{table:maskrcnn}. 
We observe that the model trained on images synthesized by the GIS network significantly outperforms the one trained only on real data, and marginally outperform the highly-tuned data from \cite{Alhaija2017BMVC}. This clearly indicates that our model does not only learn to imitate the training data, but also the adversarial loss can contribute in make the resulting appearance more realistic and, therefore, more effective in training.

\subsection{Generalization and Ablation Study}

A key feature of our GIS framework is that it learns a mapping from any geometry to natural images and is not limited to a specific set of objects or shapes. In the following sections, we present an extensive experimental study to demonstrate that our model learns a generic image formation function and does not overfit or limit itself only to objects of certain geometry and material. 

\begin{figure}[t!]
	\centering
	\begin{subfigure}[b]{0.48\textwidth}
		\includegraphics[width=\textwidth]{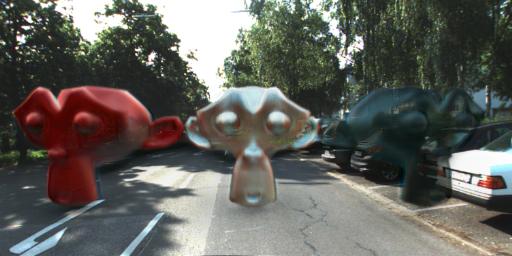}
		\caption{}
		\label{fig:material_monkey}
	\end{subfigure}
	\begin{subfigure}[b]{0.48\textwidth}
		\includegraphics[width=\textwidth]{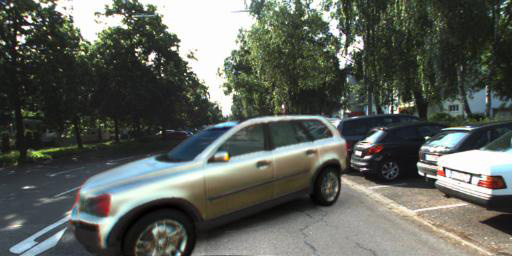}
		\caption{}
		\label{fig:material_chrome}
	\end{subfigure}
	\caption{(a) GIS output for a monkey model with material labels: car paint, chrome and glass. (b) GIS output for a car with material label chrome.}\label{fig:material}
\end{figure}

\begin{figure}[t!]
\centering
\includegraphics[width=\textwidth]{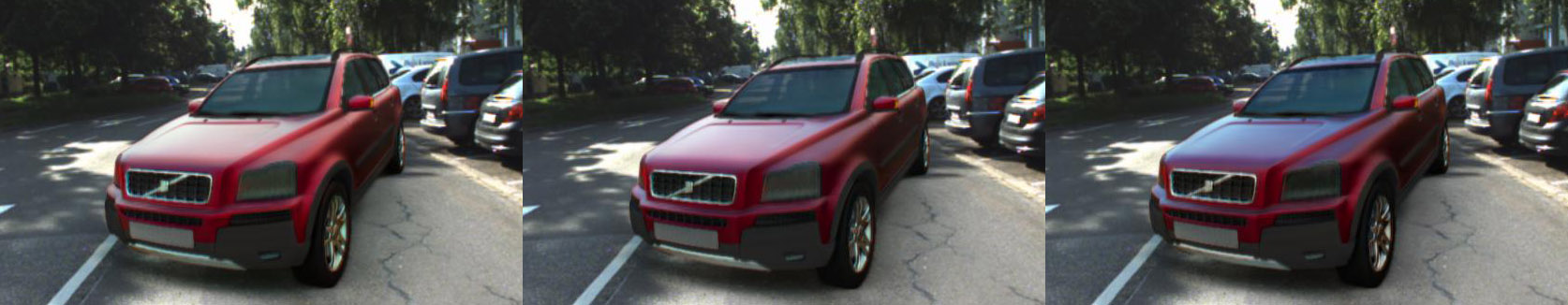}
\caption{Three diverse outputs obtained from GIS on the KITTI-360 dataset. Note how the renderings vary in lighting conditions and reflections.}
\label{fig:diversity}
\end{figure}

To show generalization ability, we present the network with two tasks: (i) synthesize seen objects with material combinations never seen before, and (ii) synthesize learned materials on new, unseen, geometries. In \figref{fig:material_monkey} we show the results of our model applied to the "Monkey" model from blender with different material labels applied to it. Our results clearly demonstrate that the material properties have been learned by the network independently from the geometry. In \figref{fig:material_chrome} we replace the car paint with the chrome material previously seen in the training data only on the car wheel rims. The resulting image looks realistic, demonstrating that the material properties learned from one part of the model can be transferred to other, geometrically different, parts by simply changing the material label. Using the diversity loss, described in Section \ref{subsec:method_diversity}, our GIS model produces 9 different possible images from the same input. The results in \figref{fig:diversity} show how the network can learn different lighting conditions (direct light, cloudy etc.) without providing any explicit lighting information.

\begin{figure}[t!]
    \centering
    \begin{subfigure}[b]{0.40\textwidth}
        \includegraphics[width=\textwidth]{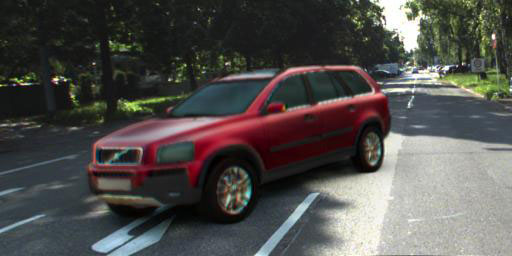}
        \caption{{Normal, depth, material, mask}}
        \label{fig:input_all}
    \end{subfigure}
    \begin{subfigure}[b]{0.40\textwidth}
        \includegraphics[width=\textwidth]{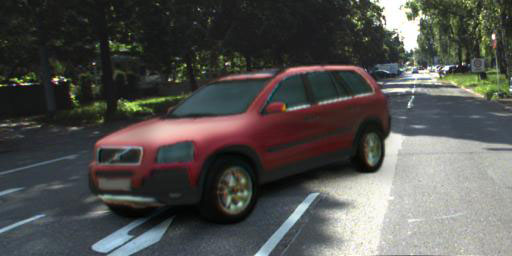}
        \caption{All inputs except normals}
        \label{fig:input_nonormal}
    \end{subfigure}
\\
    \begin{subfigure}[b]{0.40\textwidth}
        \includegraphics[width=\textwidth]{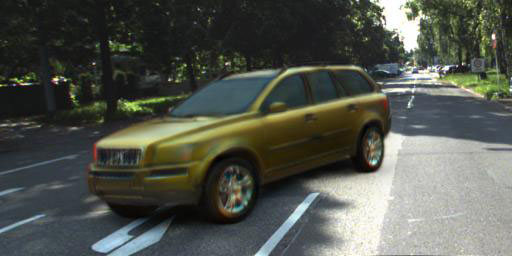}
        \caption{All inputs except material ids}
        \label{fig:input_nomaterial}
    \end{subfigure}
    \begin{subfigure}[b]{0.40\textwidth}
        \includegraphics[width=\textwidth]{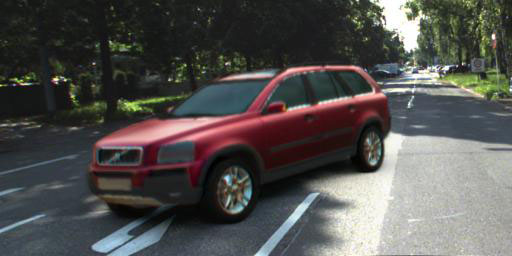}
        \caption{All inputs except depth}
        \label{fig:input_nodepth}
    \end{subfigure}
    \caption{Output of GIS using various types of inputs. Note that GIS with all four inputs, or all inputs except depth, synthesizes realistic images. }
    
    \label{fig:inputs}
\end{figure}

To better understand the importance of different input modalities, we perform an ablation study where we train the GIS model from scratch using all inputs excluding one at a time. We qualitatively compare the results in \figref{fig:inputs}. When normals are not used for training, the output images become smooth and lack fine geometric details. Excluding depth maps from the input, on the other hand, leads to no noticeable difference. We hypothesize that this is due to the fact that most of the shading of the object can be modeled based on the local geometry cues that are expressed well in the normals, but little difference in appearance relates to the absolute depth of an object.  In contrast, removing the material segmentation results in blurry images.

\begin{table}[H]
\begin{center}
    
\begin{tabular}{l|cccccccc}
\hline
Dataset  & IoU 50\% & AP \\
\hline
Real KITTI-360  		& 58.80\%	& 31.92\% \\
Augmented KITTI-360		& 66.68\%	& 37.88\% \\
GIS (ours)		& 67.74\%	& 38.69\% \\
\hline
\end{tabular}
\end{center}
\caption{Accuracy of Mask R-CNN when trained with real, augmented or GIS generated images.}
\label{table:maskrcnn}
\end{table}

\subsection{Novel view synthesis on Linemod dataset}
The Linemod dataset was introduced by Hinterstoisser \etal\cite{hinterstoisser2012accv} for evaluating 6D object pose estimation algorithms. 
This dataset contains real images of multiple known objects, each of them annotated with a 6 degree-of-freedom pose. It provides 3D CAD models of all the objects in the dataset for which we annotated the materials present on each CAD model with a material label. Hence, using the 6D pose and the CAD model, we can obtain the normal map ($N$), material segmentation ($S$) and depth map ($D$) of objects. 

The objective of this experiment is to use real images as target training data for our network with the corresponding geometric information ($N,S,D$) as inputs. Unlike the KITTI-360 dataset, where the target data is acquired using a manually designed rendering framework, the availability of real images as target data in this case allows the network to model real world images and their statistics directly. We use the objects ${Ape, Can, Cat, Duck, Eggbox, Holepuncher}$ each with 1200 images and their pose annotations. We use 600 images of each object for training and the rest for testing. 

\begin{figure}[t!]
\centering
\includegraphics[width=\textwidth]{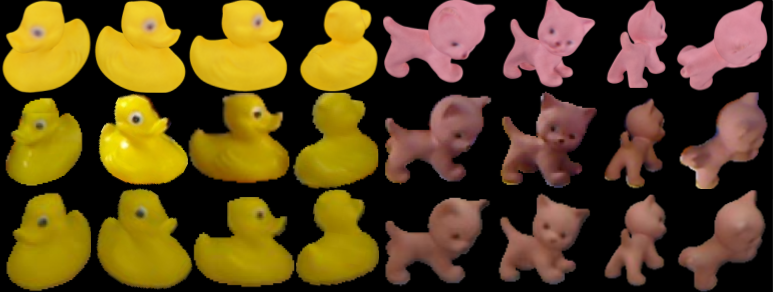}
\caption{Top row contains rendered images. Middle row contains real images in similar poses. Bottom row contains images synthesized by our network. The middle row images are not seen during training phase. GIS is still able to synthesize novel views of objects realistically. }
\label{fig:linemod_fig}
\end{figure}

Due to the generalizability of our method, we can use the trained GIS network to generate new images of the objects in previously unseen poses. 
To demonstrate the efficacy of the images produced by our GIS network for training, we compare them to a training set generated using the traditional OpenGL rendering engine.
To this end, we use the two kinds of datasets to train the Mask R-CNN. First, we crop the rendered images and place them at random locations on NYU dataset \cite{Silberman2012ECCV} images. We repeat the same process for object images generated by our network. To evaluate the performance of Mask R-CNN, we test it on Linemod-Occluded dataset, proposed by Michel \etal \cite{michelcvpr2017} (note that we do not use this data while training our GIS network). We observe that the Mask R-CNN trained with rendered images performs at an average accuracy of $21.1\%$ for all objects. On the other hand, the Mask R-CNN trained with our synthesized images performs at an average accuracy of $68.4\%$. This clearly indicates that the images synthesized by our network are highly realistic and are useful for training other deep networks which cannot be achieved with rendered data. Qualitatively, our GIS generated images appear more similar to real images as shown in Fig. \ref{fig:linemod_fig}. 

%% file: sec05_conclusion.tex
\section{Conclusion}
In this we work, we have proposed GIS, a deep neural network which is able to learn to synthesize realistic objects by leveraging semantic and geometric scene information. Through various experiments we have demonstrated the generalization performance of our GIS framework with respect to varying geometry, semantics and materials. Further, we have provided empirical evidence that the images synthesized by GIS are realistic enough to train the state-of-the-art instance segmentation method Mask R-CNN, and improve its accuracy on car instance segmentation with respect to a baseline model trained on non-augmented images from the same dataset. We believe that our approach opens new avenues towards ultimately reaching the goal of photo-realistic image synthesis using deep neural networks. 

\section{Acknowledgments}
This project has received funding from the European Research Council (ERC) under the European Unions Horizon 2020 programme (grant No. 647769) and by the Heidelberg Collaboratory for Image Processing (HCI).